\pdfoutput=1

\documentclass[11pt]{article}

\usepackage[]{coling}

\usepackage{times}
\usepackage{latexsym}

\usepackage[T1]{fontenc}

\usepackage[utf8]{inputenc}

\usepackage{microtype}

\usepackage{inconsolata}

\usepackage{graphicx}

\usepackage{microtype}
\usepackage{hyperref}
\usepackage{url}
\usepackage{booktabs}

\usepackage{graphicx} 
\usepackage{amsmath}
\usepackage{amsfonts}
\usepackage{amsthm}

\usepackage{multirow} 

\usepackage[many]{tcolorbox}	
     
\tcbset{
    sharp corners,
    colback = white,
}   
\newtcolorbox{boxA}{
    rounded corners,
    arc = 5pt 
}

\usepackage{colortbl}  
\usepackage{xcolor}
\usepackage{array}
\usepackage{arydshln}
\definecolor{gg}{HTML}{e2f0cb}

\title{Rethinking Kullback-Leibler Divergence in \\ Knowledge Distillation for Large Language Models}

\newcommand*{\affmark}[1][*]{\textsuperscript{#1}}
\usepackage{fdsymbol}

\author{
Taiqiang Wu\affmark[$\diamondsuit$] \
Chaofan Tao\affmark[$\diamondsuit$] \ Jiahao Wang\affmark[$\diamondsuit$] \
Runming Yang\affmark[$\clubsuit$] \\ \textbf{Zhe Zhao}\affmark[$\spadesuit$] \ \textbf{Ngai Wong}\affmark[$\diamondsuit$]
\\
\affmark[$\diamondsuit$]The University of Hong Kong \ \affmark[$\clubsuit$]Tsinghua University \
\affmark[$\spadesuit$]Tencent AI Lab
\\
{\tt \{takiwu,cftao,jiahao.wang\}@connect.hku.hk} \ {\tt nwong@eee.hku.hk}
}

\begin{document}
\maketitle
\begin{abstract}
Kullback-Leiber divergence has been widely used in Knowledge Distillation~(KD) to compress Large Language Models~(LLMs).
Contrary to prior assertions that reverse Kullback-Leibler~(RKL) divergence is mode-seeking and thus preferable over the mean-seeking forward Kullback-Leibler~(FKL) divergence, this study empirically and theoretically demonstrates that neither mode-seeking nor mean-seeking properties manifest in KD for LLMs.
Instead, RKL and FKL are found to share the same optimization objective and both converge after a sufficient number of epochs.
However, due to practical constraints, LLMs are seldom trained for such an extensive number of epochs.
Meanwhile, we further find that RKL focuses on the tail part of the distributions, while FKL focuses on the head part at the beginning epochs.
Consequently, we propose a simple yet effective Adaptive Kullback-Leiber~(AKL) divergence method, which adaptively allocates weights to combine FKL and RKL. 
Metric-based and GPT-4-based evaluations demonstrate that the proposed AKL outperforms the baselines across various tasks and improves the diversity and quality of generated responses.
Codes are available at \href{https://github.com/wutaiqiang/LLM_KD_AKL}{github}.
\end{abstract}

\section{Introduction}

Large Language Models~(LLMs), such as GPT-4 \citep{DBLP:journals/corr/abs-2303-08774}, OPT \citep{DBLP:journals/corr/abs-2205-01068}, and LLaMA  \citep{DBLP:journals/corr/abs-2302-13971}, have achieved great success in various Natural Language Process~(NLP) tasks.
Such strong capabilities are often attributed to the increased scale of training data and model size, such as the 13 billion parameters in LLaMA \citep{DBLP:journals/corr/abs-2302-13971}.
However, large model size also brings expensive computing costs and high latency, which limit the deployment for real-world applications.
Therefore, it is crucial to compress the model size while keeping the performance as much as possible.
Knowledge Distillation~(KD, \citep{DBLP:journals/corr/HintonVD15}), which trains a compact student model by mimicking the behaviors of the teacher model, has been a promising method to compress the Language Models~(LMs) \citep{DBLP:journals/corr/abs-2305-09098}.
Hence, how to apply KD methods appropriately to compress LLMs has been a hot topic recently.

The vanilla KD methods \citep{DBLP:journals/corr/HintonVD15} try to train a better student model by aligning the logits $q_{\theta}$ with logits $p$ from the teacher model via forward KL~(FKL) divergence $FKL(p,q_{\theta})=\sum_{z} p(z)\log(p(z)/q_{\theta}(z))$.
Meanwhile, the reverse KL~(RKL) divergence is defined as $RKL(p,q_{\theta})=\sum_{z} q_{\theta}(z)\log(q_{\theta}(z)/p(z))$ and typically $RKL\neq FKL$ due to the asymmetry of KL divergence.\footnote{In this paper, we denote FKL divergence and RKL divergence as FKL and RKL, respectively.}.
For continuous $p$ and learnable $q_{\theta}$, FKL causes the mean-seeking behavior where $q_{\theta}$ tends to cover the main mass of $p$ simultaneously,
while RKL causes the mode-seeking behavior where $q_{\theta}$ tends to select one mode \citep{DBLP:books/daglib/0023091,DBLP:journals/jmlr/0001SLKMW22,wang2023beyond}.
Therefore, recent methods \citep{DBLP:journals/corr/abs-2306-08543, agarwal122024policy, kim2024promptkd} argue that RKL is more suitable than FKL to align the logits from LLMs since RKL avoids learning too many long-tail variants.

However, such mean-seeking and mode-seeking behaviors rely on \textbf{two assumptions}: 1) the student distribution $q$ is unimodal such as the Gaussian distribution, and 2) teacher distribution $p$ and student distribution $q$ are continuous.
Intuitively, these two conditions \textbf{do not hold} in the KD for LLMs, where $p$ and $q$ are discrete and $q$ is not guaranteed to be unimodal.
In this paper, we prove the aforementioned conclusion empirically and theoretically, and further show that both FKL and RKL converge to the same objective after a sufficient number of epochs~(more than 50 epochs in our demonstration experiments).
In practice, we do not train for such an extensive number of epochs, such as 10 epochs in \citet{DBLP:journals/corr/abs-2306-08543}.
Meanwhile, we find that RKL focuses on the tail part of the teacher distribution $p$, while FKL focuses on the head part at the beginning epochs.
Based on this observation, we propose a novel Adaptive Kullback-Leiber~(AKL) divergence, aiming to better align the distributions.
The key insight is to adaptively allocate the weights to combine the FKL and RKL based on $p$ and $q$ distributions.

We evaluate AKL on various popular benchmarks.
The experimental results demonstrate that AKL outperforms all the baseline methods.
Besides the Rouge-L scores, we also employ the GPT-4 \citep{DBLP:journals/corr/abs-2303-08774} to score the generated responses on diversity and quality under different seeds.
The GPT-4 scores show that AKL can improve the diversity and quality of generated responses. 
Our contributions can be concluded as follows:
\begin{itemize}
    \item We demonstrate that the mean-seeking behavior of FKL and mode-seeking behavior of RKL do not hold in the KD for LLMs.
    Instead, both FKL and RKL converge to the same objective after a sufficient number of epochs.
    \item We find that FKL focuses on the head part and RKL focuses on the tail part at the beginning epochs, and thus propose a novel AKL to align the distributions better under typically limited epochs.
    \item We demonstrate the effectiveness of AKL via results on various benchmarks.
    Moreover, GPT-4 scores indicate that AKL improves the diversity and quality of the generated responses.
\end{itemize}
\section{Related Work}

\subsection{KD for LLM Compression}

Typically, KD methods for LLM can be categorized into 1) black-box KD, where only the teacher predictions are available, and 2) white-box KD, where more output from the teacher model can be employed.
The Black-Box KD methods \citep{kim2016sequence,taori2023stanford, chiang2023vicuna} employ the responses of teacher models to improve the training data for student models.
Meanwhile, white-box KD methods align the logits from teacher models and student models during the token generation process, through token-scale forward KL  \citep{DBLP:conf/nips/KimLLHCSC23} or reverse KL  \citep{DBLP:journals/corr/abs-2306-08543, agarwal122024policy, kim2024promptkd}.
In this paper, we first revisit the forward KL and reverse KL in the KD for LLMs and then propose a novel AKL.

\subsection{Forward KL and Reverse KL}

Based on the analysis in \citet{minka2005divergence}, reverse KL is zero forcing because $p=0$ forces $q=0$, and the forward KL tends to model the main mass as requiring $q>0$ whenever $p>0$.
Due to the zero forcing behavior, reverse KL is widely used in domain adaption tasks \citep{nguyen2021kl} and reinforcement learning \citep{czarnecki2019distilling}.
Meanwhile, forward KL is firstly employed in knowledge distillation by \citet{DBLP:journals/corr/HintonVD15}.
After that, \citet{lee2023self} simply adds the forward KL and reverse KL for KD.
\citet{amara2022bd} proposes BD-KD, a weighted sum of forward KL and reverse KL for online KD tasks, where the weights are either 1 or $\lambda$~(hyper-parameter, $\lambda>1$) based on the entropy of student and teacher.
Compared to BD-KD, our proposed AKL assigns adaptive weights based on the head and tail parts rather than the entropy of the whole distribution.


\section{Preliminary and Rethinking}

\subsection{Background}
Conditional language generation tasks aim to generate a response $\mathbf{y}=\{y_t\}^T_{t=1}$ based on the input $\mathbf{x}$, where $y_t \in \{Y_1,Y_2,...,Y_V\}$ and $V$ is the vocabulary size.
Traditional FKL aims to align the token generation distributions, which can be decomposed into a step-wise KL divergence \citep{wen2023f} as follows:
\begin{equation}
\begin{aligned}
\mathcal{J}_{FKL} &=\sum\limits_{t=1}^{T} FKL(p(y_t|\mathbf{y}_{<t}),q_{\theta}(y_t|\mathbf{y}_{<t})) \\
&=\sum\limits_{t=1}^{T} 
\sum\limits_{j=1}^{V}
p(Y_j|\mathbf{y}_{<t}) \log(\frac{p(Y_j|\mathbf{y}_{<t})}{q_{\theta}(Y_j|\mathbf{y}_{<t})}).
\end{aligned}
\end{equation}

Meanwhile, we can replace the FKL with RKL, and thus get:
\begin{equation}
\begin{aligned}
\mathcal{J}_{RKL} &=\sum\limits_{t=1}^{T} RKL(p(y_t|\mathbf{y}_{<t}),q_{\theta}(y_t|\mathbf{y}_{<t})) \\
&=\sum\limits_{t=1}^{T} 
\sum\limits_{j=1}^{V}
q_{\theta}(Y_j|\mathbf{y}_{<t})
\log(\frac{q_{\theta}(Y_j|\mathbf{y}_{<t})}{p(Y_j|\mathbf{y}_{<t})}).
\end{aligned}
\end{equation}

\begin{figure}[t]
\begin{center}
\includegraphics[width=0.9\linewidth]{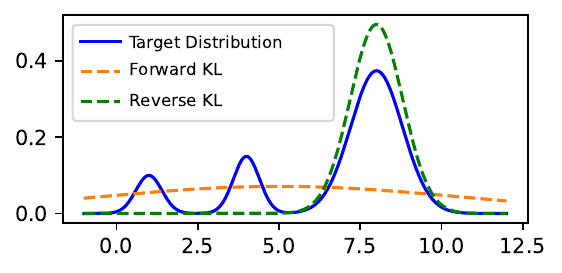}
\end{center}
\caption{The toy example in \citet{DBLP:journals/corr/abs-2306-08543}, where they fit a Gaussian mixture~(distribution of teacher) with a Gaussian distribution~(distribution of student) using FKL and RKL.
}
\label{minillm_toy}
\end{figure}

Recent methods \citep{DBLP:journals/corr/abs-2306-08543, agarwal122024policy, kim2024promptkd} argue that RKL is more suitable than FKL because FKL leads to the mean-seeking behavior and RKL leads to the mode-seeking behavior.
As shown in Figure \ref{minillm_toy}, mean-seeking means the $q_{\theta}$ tries to cover the main mass~(the mean of three peaks) of $p$ simultaneously.
Meanwhile, mode-seeking means $q_{\theta}$ assigns high probabilities to the main mass~(largest peak).

\subsection{Rethinking FKL and RKL}

\paragraph{Empirical Analysis.}
However, the mean-seeking behavior of FKL and the mode-seeking behavior of RKL relies on two assumptions: 1) student distribution $q_{\theta}$ follows the Gaussian distribution, and 2) both teacher distribution $p$ and student distribution $q_{\theta}$ are continuous.
Intuitively, since $q_{\theta}$ follows the Gaussian distribution, then the learned distribution needs to select where to cover most, either the main mass~(FKL) or one peak~(RKL). 

\begin{figure*}[t]
\begin{center}
\includegraphics[width=\linewidth]{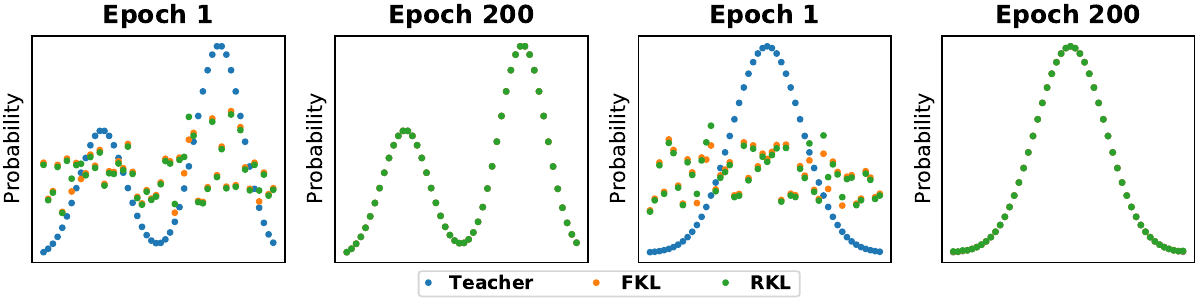}
\end{center}
\caption{The convergence of FKL and RKL on toy data under epoch 1 and epoch 200.
The initial distribution $q$ is the same for FKL and RKL.
After 200 epochs, both FKL and RKL can converge to the target distribution well regardless of the shape of $p$.
}
\label{epoch_1_200}
\end{figure*}

In the KD for LLMs, these two assumptions do not hold anymore.
On the one side, student distribution $q_{\theta}$ is calculated by the softmax function and does not follow the standard Gaussian distribution.
On the other side, both teacher distribution $p$ and student distribution $q_{\theta}$ are discrete and defined on vocabulary set $\{Y_1,Y_2,...,Y_V\}$.
Figure \ref{epoch_1_200} shows the results on toy examples under the setting of KD for LLMs.
For FKL and RKL, we employ the same initialized $q_{\theta}$ and train for 200 epochs.
We calculate the FKL/RKL loss and update the $q_{\theta}$ via the Adam optimizer \citep{diederik2014adam} without weight decay.
The learning rate is 0.1.
Both FKL and RKL converge to $p$ perfectly after optimizing for 200 epochs regardless of the shape of $p$.

\paragraph{Theoretical Analysis.}
We also demonstrate the phenomena theoretically.
Since the $\mathcal{J}_{FKL}$ and $\mathcal{J}_{RKL}$ are additive considering generated tokens, we can analyze the convergences at each token separately. 
Let $(z^p_1, z^p_2, z^p_3,..., z^p_V)$ and $(z^q_1, z^q_2, z^q_3,..., z^q_V)$ denote the output at step $t$ before the final softmax operation from the teacher model and the student model, respectively.
We have
\begin{equation}
    p(Y_j|\mathbf{y}_{<t}) = 
    \frac{\exp(z^p_j)}{\sum\limits_{k=1}^{V}\exp(z^p_k)}
    ; 
\end{equation}
\begin{equation}
    q_{\theta}(Y_j|\mathbf{y}_{<t}) = \frac{\exp(z^q_j)}{\sum\limits_{k=1}^{V}\exp(z^q_k)}.
\end{equation}
Then the gradient for $z^q_j$ under FKL and RKL can be calculated by the chain rule as follows:
\begin{equation}
\frac{\partial FKL(p,q_{\theta})}{\partial z^q_j} = q_{\theta}(Y_j|\mathbf{y}_{<t}) - p(Y_j|\mathbf{y}_{<t})
\end{equation}
\begin{equation}
\begin{aligned}
\frac{\partial RKL(p,q_{\theta})}{\partial z^q_j} = q_{\theta}(Y_j|\mathbf{y}_{<t}) (\log(\frac{ q_{\theta}(Y_j|\mathbf{y}_{<t})}{p(Y_j|\mathbf{y}_{<t})}) \\ -RKL(p,q_{\theta})).
\end{aligned}
\end{equation}
Considering the converge condition:
\begin{equation}
\frac{\partial F(R)KL(p,q_{\theta})}{\partial z^q_j} = 0 \ \forall j \in \{1,2,3,...,V\},
\end{equation}
we can infer that for both FKL and RKL, the sufficient and necessary condition for converge is
\begin{equation}
q_{\theta}(Y_j|\mathbf{y}_{<t}) = p(Y_j|\mathbf{y}_{<t}) \ \forall j \in \{1,2,3,...,V\}.
\end{equation}

Please refer to Appendix \ref{FRKL_proof} for the detailed proof.
In summary, RKL and FKL share the same optimization objective, forcing the student model to generate the same logits as the teacher model.

\paragraph{Deeper Insights.} \ 
FKL and RKL are special cases of f-divergence \citep{sason2016f}.
The f-divergence is defined as:
\begin{equation}
D_f(P||Q)= \int f(\frac{dP}{dQ}) dQ,
\end{equation}
where $f$ is convex and $f(1)=0$.
Following the Jensen's inequality:
\begin{equation}
D_f(P||Q)= \int f(\frac{dP}{dQ}) dQ \ge  f(\int \frac{dP}{dQ}dQ) = 0.
\label{none_neg}
\end{equation}

For FKL $f(x)=x\log(x)$, and for RKL $f(x)=-\log(x)$.
Then we can revisit the mode-seeking and mean-seeking behaviors in the view of f-divergence:
\begin{itemize}
    \item Taking $f(x)=x\log(x)$, we have $\lim\limits_{x \to +\infty}f(x)=+\infty$, so the FKL avoids $\frac{p(z)}{q(z)}$ goes to $+\infty$. Since $0<p(z)<1$ and $0<q(z)<1$, $q(z)$ should not be too small especially when $p(z)$ is large.
    Therefore, $q$ would try to cover~(assign larger value) as many peaks of $p$ as possible, leading to the mean-seeking behavior of FKL.
    \item Taking $f(x)=-\log(x)$, we have $\lim\limits_{x \to +\infty}f(x)=-\infty$ and $\lim\limits_{x \to 0^{+}}f(x)=+\infty$.
    According to Equation \ref{none_neg}, RKL is nonnegative and thus only avoids $\frac{p(z)}{q(z)}$ go to $0^{+}$, which means $q(z)$ should not be too large~($q(z) \to 1$) when $p(z)$ is small~($p(z) \to 0$).
    Therefore, $q$ would cover one mode of $p$ when $q$ follows the Gaussian distribution, which is also known as the mode-seeking behavior of RKL.
    
\end{itemize}

\paragraph{Difference between FKL and RKL.}
In this paper, we demonstrate that the mean-seeking behavior of FKL and the mode-seeking behavior of RKL do not hold in the KD for LLMs empirically and theoretically.
Instead, FKL and RKL share the same optimizing objective, that is, $q_{\theta}$ from the student is the same as $p$ from the teacher. 
We thus ask:
\textit{What is the difference between FKL and RKL in the KD for LLMs?}

The answer is the optimization process.
Considering $FKL(p,q_{\theta})=\sum p(z)\log\frac{p(z)}{q_{\theta}(z)}$, larger $p(z)$ means a larger weight in total loss and also more likely to generate a larger $\log\frac{p(z)}{q_{\theta}(z)}$.
Hence, fitting the area with larger $p(z)$~(a.k.a. the head part) is the priority of FKL.
Similarly, $RKL(p,q_{\theta})$ is defined as $\sum q_{\theta}(z)\log\frac{q_{\theta}(z)}{p(z)}$.
When $p(z)$ is smaller, $\frac{q_{\theta}(z)}{p(z)}$ is easier to be $+\infty$, leading to a larger loss.
Therefore, fitting the area with smaller $p(\cdot)$~(a.k.a. the tail part) is the priority of RKL.

\begin{figure*}[t]
\begin{center}
\includegraphics[width=\linewidth]{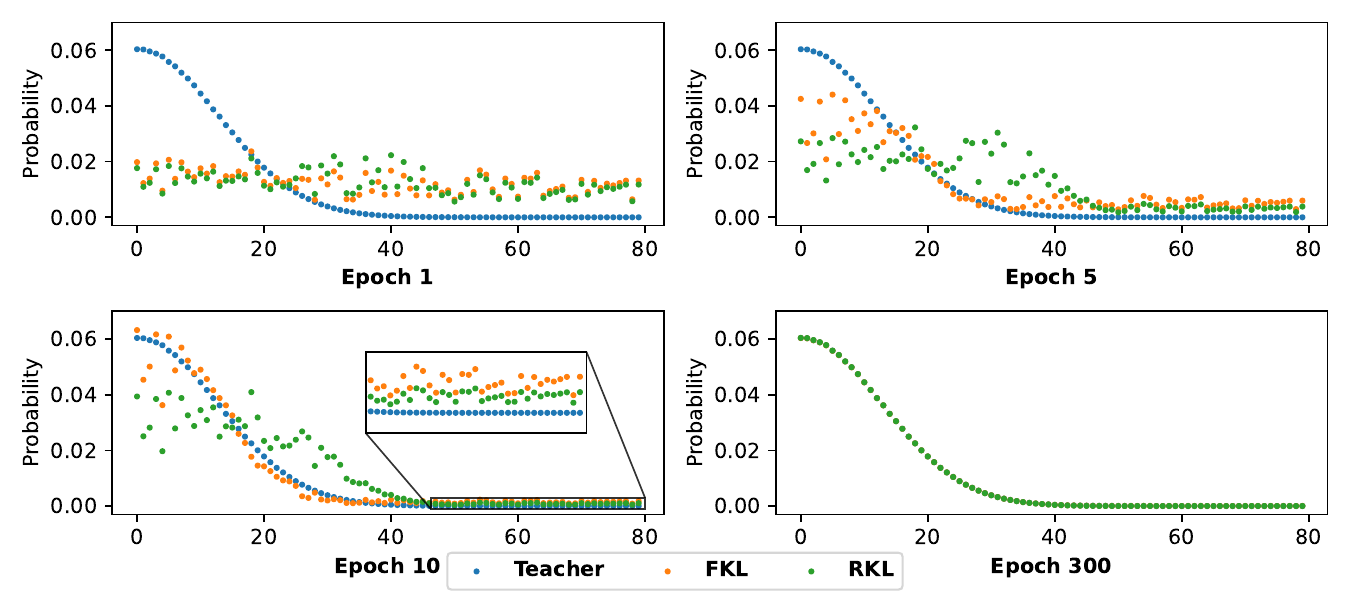}
\end{center}
\caption{The distributions at various epochs for FKL and RKL on toy data (long-tail), where the teacher distribution and initial student distribution are the same.
We can find that FKL focuses on the \textit{head} part and RKL on the \textit{tail} part at the beginning epochs, and both converge finally.
}
\label{converg}
\end{figure*}

Figure \ref{converg} shows the results on the long-tail toy data.
For RKL and FKL, we employ the same teacher and initial student distribution.
After training for 250 epochs, both FKL and RKL converge as mentioned before.
However, we can find that RKL fits better at the tail part and FKL fits better at the head part at beginning epochs.

\paragraph{Summary}
Deeper insights on FKL and RKL in KD for LLMs can be concluded as follows.

\begin{boxA}
\textit{In the KD for LLMs, the mean-seeking and mode-seeking behaviors do not hold for forward KL~(FKL) and reverse KL~(RKL), respectively.
Instead, they share the same optimization objective.
Meanwhile, FKL focuses on the head part and RKL focuses on the tail part at the beginning.}
\end{boxA}


\section{Methodology}

\subsection{Motivation}
In Figure \ref{converg}, we can find that both FKL and RKL converge to the teacher distribution after 300 epochs.
However, when performing experiments on LLMs, we \textbf{do not train for such an extensive number of epochs due to unaffordable costs}, especially on the large corpus.
For example, \citet{DBLP:journals/corr/abs-2306-08543} trains only 20 epochs for models with less than 1.3B parameters and 10 epochs for others.
Meanwhile, under such a small epoch, FKL and RKL focus on the head and tail parts, respectively.
Both FKL and RKL fail to align the logit distributions well.
To tackle it, one intuition is to combine both the FKL and the RKL.
However, how to assign the loss weights to balance FKL and RKL becomes the next challenge.

\subsection{Adaptive Kullback-Leiber Divergence}

In this paper, we propose a novel Adaptive Kullback-Leiber~(AKL) divergence method, which adaptively allocates the weights to combine FKL and RKL.
The key idea is to assign larger weights to FKL if the gap on the head part is larger than the tail part, and vice versa.
Specifically, for the logit distribution $p(y_t|\mathbf{y}_{<t}) \in \mathbb{R}^{V}$ from the teacher model and $q_{\theta}(y_t|\mathbf{y}_{<t})\in \mathbb{R}^{V}$ from the student model at token $t$, we first define the indicate mask $M\in \mathbb{R}^{V}$ as follows:
\begin{equation}
    M[i] = 
\begin{cases}
1 \ \ when\ Y_i \ belongs \ to \ head \ part \\
0 \ \ otherwise.
\end{cases}
\end{equation}

In AKL, we can get one feasible $M$ by solving:
\begin{equation}
\begin{aligned}
    min \ \sum\limits_{i=0}^{V} M[i] \ \ s.t. \ \sum\limits_{j=0}^{V} M[j]p(Y_j|\mathbf{y}_{<t}) \ge \mu,
\end{aligned}
\end{equation}
where the hyper-parameter $\mu$ is the threshold such as 0.5.
We can easily solve this optimizing problem by sorting the $p(Y_j|\mathbf{y}_{<t})$.
After that, the gap $g_{head}$ on the head part and $g_{tail}$ on the tail part are calculated by:
\begin{equation}
    g_{head} = \sum\limits_{i=0}^{V} M[i]\epsilon(p(Y_i|\mathbf{y}_{<t}), q_{\theta}(Y_i|\mathbf{y}_{<t}));
\end{equation}
\begin{equation}
    g_{tail} = \sum\limits_{i=0}^{V} (1-M[i])\epsilon(p(Y_i|\mathbf{y}_{<t}), q_{\theta}(Y_i|\mathbf{y}_{<t}))
\end{equation}
where $\epsilon(\cdot)$ is the function to define the point-wise gap.
Following the idea of Jeffreys divergence~\cite{jeffreys1946invariant}, AKL is defined as
\begin{equation}
\begin{aligned}
AKL(p,q_{\theta}) = &\frac{g_{head}}{g_{head}+g_{tail}}FKL(p,q_{\theta}) \\+ &\frac{g_{tail}}{g_{head}+g_{tail}}RKL(p,q_{\theta}).
\label{equ_akl}
\end{aligned}
\end{equation}
In this way, a larger gap $g_{head}$ would bring a larger weight to FKL, and vice versa.

\section{Experiments}

\subsection{Experimental Setup}

\paragraph{Training}
For training data, we employ the instruction response dataset following \citet{DBLP:journals/corr/abs-2306-08543}, which is built from databricks-dolly-15k\footnote{
\href{https://github.com/databrickslabs/dolly/tree/master}{databrickslabs}} and contains 14k samples for training, 500 samples for validation, and 500 samples for testing.
We first fine-tune the teacher model and then distill the teacher model.
For teacher models, we select GPT-2 \citep{radford2019language} with 1.5B parameters and LLaMA \citep{DBLP:journals/corr/abs-2302-13971} with 6.7B parameters.
Meanwhile, the students are GPT-2 with 120M parameters and TinyLLaMA \citep{zhang2024tinyllama} with 1.1B parameters\footnote{\href{https://huggingface.co/TinyLlama/TinyLlama-1.1B-intermediate-step-1195k-token-2.5T}{TinyLLaMA 1.1B}}, respectively.

\begin{table*}[t]
\centering
\resizebox{0.9\linewidth}{!}
{%
\begin{tabular}{lcccccc}
\toprule
\multirow{3}{*}{\bf Method}  
& \multicolumn{3}{c}{\bf GPT2 1.5B $\rightarrow$ GPT2 120M}        & \multicolumn{3}{c}{\bf LLaMA 6.7B $\rightarrow$ TinyLLaMA 1.1B} \\ \cmidrule(lr){2-4} \cmidrule(lr){5-7}
& Dolly & S-NI & UnNI & Dolly & S-NI & UnNI \\
& (500) & (1694) & (10000) & (500) & (1694) & (10000) \\
\midrule
Teacher    & 26.98$\pm$0.27                                            & 27.25$\pm$0.16                                           & 31.61$\pm$0.16                                             & 26.73$\pm$0.65                                            & 32.75$\pm$0.24                                           & 34.61$\pm$0.08                                             \\  
SFT & 23.01$\pm$0.18                                            & 16.48$\pm$0.28                                           & 18.43$\pm$0.09                                             & 22.05$\pm$0.38                                            & 27.79$\pm$0.20                                           & 25.96$\pm$0.13                                             \\
\midrule
SeqKD      & 23.30$\pm$0.38                                            & 16.35$\pm$0.17                                           & 18.51$\pm$0.04                                             & 22.67$\pm$0.57                                            & 26.97$\pm$0.32                                           & 
27.35$\pm$0.06 \\
FKL & 23.46$\pm$0.56                                            & 16.63$\pm$0.48                                           & 19.27$\pm$0.06                                             & 22.24$\pm$0.38                                            & 28.07$\pm$0.41                                           & 26.93$\pm$0.09                                             \\
RKL        & 22.62$\pm$0.34                                            & 17.88$\pm$0.17                                           & 19.35$\pm$0.06                                             & 23.95$\pm$0.53                                            & 28.90$\pm$0.41                                           & 27.89$\pm$0.08                                             \\
SKL & 23.47$\pm$0.49 & 16.51$\pm$0.29 & 	18.46$\pm$0.08 & 23.29$\pm$0.54 &	29.89$\pm$0.18 	&29.15$\pm$0.20  \\
SRKL & 23.25$\pm$0.22 & 17.54$\pm$0.30 & 	19.31$\pm$0.10 & 22.09$\pm$0.22 &	29.60$\pm$0.38 	& 28.81$\pm$0.16  \\
FKL+RKL    & 23.36$\pm$0.53                                            & 17.83$\pm$0.23                                           & 20.37$\pm$0.08                                             & 24.08$\pm$0.51                                            & 30.98$\pm$0.31                                           & 30.48$\pm$0.10                                             \\
\rowcolor{gg}{AKL~(Ours)} & \textbf{23.88$\pm$0.46}                                   & \textbf{19.15$\pm$0.21}                                  & \textbf{21.97$\pm$0.13}                                    & \textbf{24.40$\pm$0.42}                                   & \textbf{31.37$\pm$0.23}                                  & \textbf{31.05$\pm$0.17}                                    \\ \bottomrule
\end{tabular}
}
\caption{The results of Rouge-L scores under different distillation settings.
Our proposed AKL outperforms all baselines without any extra parameters, especially in the larger UnNI dataset.}
\label{main_res}

\end{table*}

In AKL, we set $\mu$ as 0.5 and the gap function $\epsilon(p(z),q(z))=|p(z)-q(z)|$.
For GPT-2, we set the batch size as 32 and train for 20 epochs.
The learning rate is 5e-4. 
For TinyLLaMA, we set the batch size as 60 and train for 10 epochs.
The learning rate is 1e-5.
For all students, the maximum input length is 512.
It takes around 1h to train on 4 Nvidia A100 GPUs.

\paragraph{Evaluation}
After distillation, we evaluate the student models on Dolly~(contains 500 samples), S-NI~(contains 1694 samples), and UnNI~(contains 10000 samples).
We report the results of the Rouge-L scores for five different seeds.
More details can be found in Appendix \ref{appendix: dataset_detail}.
For the baselines, we employ the following methods:
\begin{itemize}
    \item \textbf{SFT}: directly train the student model on the dataset without distillation;
    \item \textbf{SeqKD} \citep{kim2016sequence}: train the student model on the data generated from the teacher model;
    \item \textbf{FKL/RKL}: employ forward/reverse KL during knowledge distillation;
    \item \textbf{SKL/SRKL}: employ the skew forward/reverse KL \citep{ko2024distillm} defined as the KL divergence between $p$/$q_{\theta}$ and their mix.
    We set the mixing ratio as 0.1 following \citet{ko2024distillm}.
    \item \textbf{FKL+RKL}: simply combine forward KL and reverse KL via $0.5FKL+0.5RKL$~(please refer to Appendix \ref{appendix: more_baseline} for more combinations).
\end{itemize}

In particular, we \textbf{do not} compare with MiniLLM since it pre-trains on extra corpus including OpenWebText \citep{gokaslan2019openwebtext} and training data for RoBERTa \citep{liu2019roberta}.

\subsection{Main Results}
Table \ref{main_res} reports the performance of the teacher and student models under different settings.
Some findings are summarized as follows:

\begin{itemize}
    \item Previous methods \citep{DBLP:journals/corr/abs-2306-08543, agarwal122024policy} argue that RKL is more suitable than FKL.
    However, the gap between FKL and RKL is minor as shown in Table \ref{main_res}.
    In particular, FKL gets a score of 23.46 for GPT2 on Dolly, which is larger than the 22.62 of RKL.
    \item Simply adding FKL and RKL~(a.k.a FKL+RKL) brings a better performance than FKL and RKL, which is consistent with the conclusion in previous work \citep{lee2023self}.
    Since FKL focuses on the head part and RKL focuses on the tail part, adding them would help the student learn more information.

    \item Our proposed AKL, which adaptively assigns weights based on the gaps between logits, learns a better student than all baselines \textbf{without any extra parameters}.
    In particular, on the larger dataset UnNI with 10000 samples, AKL outperforms FKL+AKL with a \textbf{p-value} less than 0.001 for both GPT 2 and TinyLLama~(i.e. 5e-9 and 9e-5, respectively).
    
\end{itemize}

Besides the rouge score, we also report the BERTscore in Appendix \ref{appendix: bert score} and the conclusion is consistent.
Moreover, please refer to Appendix \ref{appendix: akl_mu_sens} for more results on different threshold $\mu$.

\section{Extensive Analysis}

\begin{figure*}[t]
\begin{center}
\includegraphics[width=0.95\linewidth]{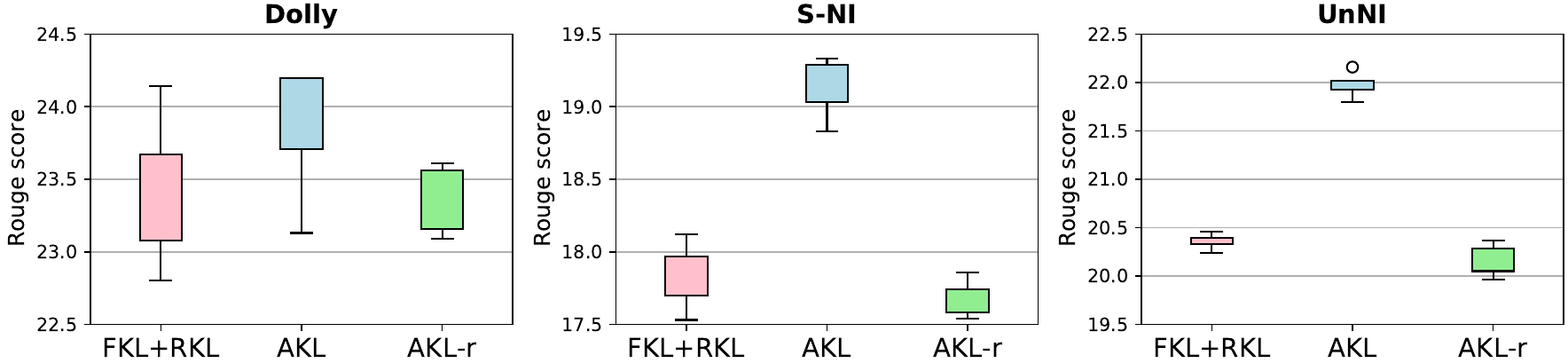}
\end{center}
\caption{The results of FKL+RKL, proposed AKL, and AKL-r on GPT 2 120M.
After flipping the loss weight, AKL-r performs worse than AKL on all three datasets.
}
\label{aba}
\end{figure*}

\subsection{Ablation: Flipping the Loss Weights}

As shown in Equation \ref{equ_akl}, we assign a larger weight to FKL if the gap on the head part is larger than the tail part.
For ablation, we flip the loss weights and then propose AKL-r, which is defined as:
\begin{equation}
\begin{aligned}
    AKL\text{-}r(p,q_{\theta}) = &\frac{g_{tail}}{g_{head}+g_{tail}}FKL(p,q_{\theta}) \\+ &\frac{g_{head}}{g_{head}+g_{tail}}RKL(p,q_{\theta}).
\end{aligned}
\end{equation}
In this way, the weight for FKL would be smaller when the gap is larger on the head part.
Figure \ref{aba} shows the results on GPT 2.
After flipping the loss weights, AKL-r performs worse than AKL and FKL+RKL, which proves the effectiveness of weight assignment in AKL.

\subsection{Computing Costs}

We report the peak GPU memory usage and training time on the experiment from LLaMA 6.7B to TinyLLaMA on the Nivida 40G A100 GPU.
As shown in Table \ref{table: training_overhead}, we can find that AKL requires slightly more GPU memory and training time.
Compared with FKL+RKL, AKL requires an extra GPU memory of 2.8G and an extra training time of 4 minutes.
However, we believe that such an extra computing overhead is acceptable since a better compact student would \textbf{save more time} on downstream tasks.

\begin{table}[t]
\centering
\resizebox{0.95\columnwidth}{!}
{%
\centering
\begin{tabular}{lcc}
\toprule
\textbf{Methods} & \textbf{GPU Memory} & \textbf{Training Time} \\
\midrule
SFT & 31.2G & 60min \\
FKL & 35.6G & 70min \\
RKL & 35.6G & 70min \\
SKL & 36.5G & 80min \\
SRKL & 37.5G & 86min \\
FKL+RKL & 37.5G & 76min \\
AKL~(Ours) & 40.3G & 80min \\
\bottomrule
\end{tabular}
}
\caption{
Comparison of computing overhead.
Compared to baselines, AKL requires slightly more GPU memory and training time.
}
\label{table: training_overhead}

\end{table}

\subsection{Evaluation via GPT-4}
\label{analysisi:gpt4}
\begin{figure*}[t]
\begin{center}
\includegraphics[width=0.95\linewidth]{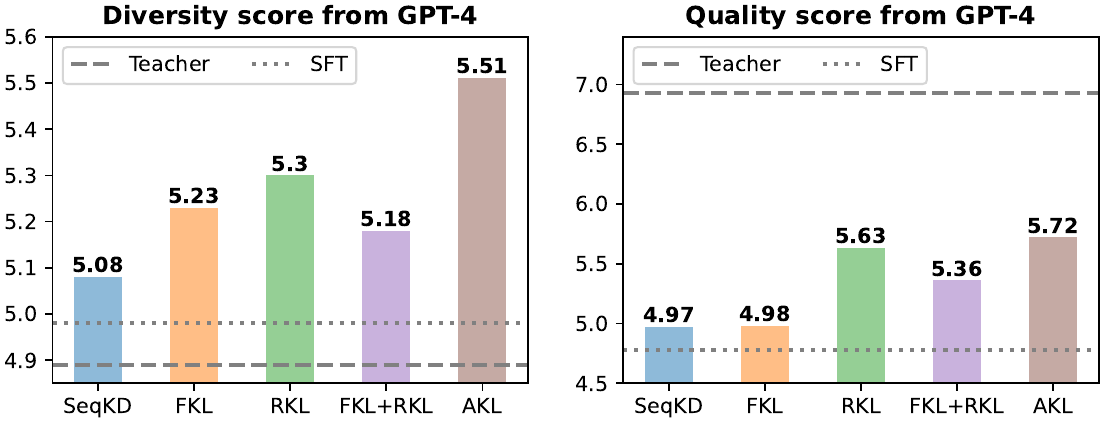}
\end{center}
\caption{The scores of diversity and quality from GPT-4~(total score: 10) on the responses from TinyLLaMA.
AKL can improve the diversity and quality of generated responses compared to the baselines.
}
\label{gpt4_scores}
\end{figure*}

We report Rouge-L scores in the aforementioned experiments to demonstrate the effectiveness of the proposed AKL.
Meanwhile, the scores from powerful LLMs such as GPT-4 are also an important assessment for generated responses.
Therefore, we design prompts and request GPT-4 to score the generated responses on \textbf{Diversity} and \textbf{Quality}.
The diversity score evaluates the ability to generate diverse answers, especially on the open question.
The quality score evaluates the ability to respond reasonably and accurately.
Please refer to Appendix \ref{appendix: gpt4score} for more details.

Figure \ref{gpt4_scores} indicates the average diversity and quality scores for the baselines and AKL.
Both FKL and RKL can improve the diversity since they mix the information~(text generation space) from teacher and student.
It is worth noting that diversity scores \textbf{do not} evaluate the correctness of the answers.
Meanwhile, FKL and RKL also improve the quality of responses by aligning with the teacher model.
In conclusion, AKL outperforms all the baselines considering diversity and quality \textbf{without} any extra parameters, demonstrating its effectiveness.

\subsection{User Study}
\label{analysisi:user}
Besides GPT-4 scores, we also perform human evaluation for the generated responses.
First, we manually filter the samples and guarantee the generated responses from these models are different. 
Then we ask 5 human experts to independently rate all the pairs for win/tie/loss. 
During evaluation, we shuffle these pairs and keep that these experts do not know the source model of each text.
As shown in Table \ref{table: human_evalu}, we can find that AKL outperforms all baselines, which is consistent with GPT-4 scores.
Specifically, AKL wins or ties all the baseline methods on more than 75\% cases considering both diversity and quality.

\begin{table}[t]
\centering
\resizebox{\columnwidth}{!}
{%
\centering
\begin{tabular}{lcc}
\toprule
\multirow{2}{*}{\textbf{Setting}} & \multicolumn{2}{c}{\textbf{Win/Tie/Loss Ratio}}  \\
\cmidrule{2-3} & Diversity & Quality  \\
\midrule
AKL vs Teacher & 36.1/44.4/19.5 & 15.9/35.6/48.5  \\
AKL vs SFT & 30.6/54.2/15.2 & 46.7/43.0/10.3  \\
AKL vs SeqKD & 33.3/48.6/18.1 & 39.3/43.0/17.7 \\
AKL vs FKL & 25.0/58.3/16.7 & 39.2/47.7/13.1 \\
AKL vs RKL & 30.6/47.2/22.2 & 27.1/48.6/24.3 \\
AKL vs FKL+RKL & 31.9/55.6/12.5 & 27.1/57.0/15.9 \\
\bottomrule
\end{tabular}
}
\caption{
Win/Tie/Loss ratios comparing AKL against baselines.
}
\label{table: human_evalu}

\end{table}

\subsection{Performance under Various Tasks}

We study the performance on various subtasks from Dolly, which contains 8 subtasks in the test set.
Appendix \ref{appendix: rouge_subdomain} shows the detailed performance of TinyLLaMA 1.1B.
FKL performs well on the Closed QA and Summary subtasks, while RKL performs well on the Brain Storming.
FKL+RKL tends to perform moderately and the scores typically lie in the range of FKL and RKL, such as 48.00 on the Classification sub-task which is between 47.01 of FKL and 51.39 of RKL.
Meanwhile, AKL adaptively assigns weights to add FKL and RKL and thus can combine the advantages well.
In particular, AKL even outperforms the teacher model on the Information Extraction and Creative Writing sub-tasks.

Moreover, we evaluate AKL for more benchmarks including Winogrande \citep{sakaguchi2021winogrande}, OpenBookQA \citep{mihaylov2018can}, BoolQ \citep{clark2019boolq}, and ARC \citep{clark2018think}.
Experimental results indicate that AKL outperforms all the baselines, demonstrating its robustness.
Please refer to Appendix \ref{appendix: more_benchmark} for more details.

\subsection{Case Study}

We further analyze the cases of generated responses from TinyLLaMA.
Please refer to Appendix \ref{appendix: case} for detailed examples and analysis.
In short, AKL can effectively distill the knowledge from the teacher model and thus improve the diversity and quality of generated responses, which is consistent with the GPT-4 scores (Section \ref{analysisi:gpt4}) and user study results (Section \ref{analysisi:user}).

\section{Conclusion}

This work first demonstrates that the mean-seeking behavior of FKL and the mode-seeking behavior of RKL do not hold in the KD for LLMs.
Instead, FKL and RKL share the same optimization objective that the student generates the same logits as the teacher, and both FKL and RKL converge after a sufficient number of epochs.
However, in practice, we train for far less epochs.
Meanwhile, we find that FKL focuses on the head part and RKL focuses on the tail part at the beginning epochs. 
Therefore, we proposed a novel AKL to add FKL and RKL adaptively.
Empirical results on various students demonstrate the effectiveness of AKL.
Besides, GPT-4 scores and human evaluation results show that AKL can improve both the diversity and the quality of generated responses.
In future work, we would consider scaling AKL to larger models with different structures and more tasks.

\section*{Acknowledgements}

We thank all anonymous reviewers for their constructive feedback on improving our paper.
This work was supported in part by the Theme-based Research Scheme (TRS) project T45-701/22-R, and in part by the General Research Fund (GRF) Project 17203224, of the Research Grants Council (RGC), Hong Kong SAR.

\section*{Limitation}

In this work, we prove that the mean-seeking and mode-seeking behaviors do not hold for forward KL~(FKL) and reverse KL~(RKL) in the KD for LLM, and propose a novel Adaptive Kullback-Leiber Divergence~(AKL).
We conduct experiments on the GPT2 and LLaMA 6.7B.
One limitation is that we have not conducted experiments on bigger language models, such as LLaMA-2 70B due to limited resources.
We leave it for future work to conduct experiments on larger models.

\section*{Ethics Statement}
This work aims to rethink the KL divergence in KD for LLM and verified on various LLMs.
However, it also inherits the social risks of generative LLMs, such as gender and representation bias \citep{lucy-bamman-2021-gender} and biases towards 'Blind' and 'Deaf'\citep{DBLP:journals/corr/abs-2206-11993}.
Fortunately, the proposed AKL can be applied to various LLMs and we encourage deploying the risk-free LLMs to reduce the potential ethical risks.


\clearpage

\appendix

\onecolumn

\section{FKL and RKL converge}
\label{FRKL_proof}

\begin{proof}
Considering the converge condition for FKL:
\begin{equation}
\frac{\partial FKL(p,q_{\theta})}{\partial z^q_j} = 0 \ \forall j \in \{1,2,3,...,V\}.
\end{equation}
We have
\begin{equation}
\frac{\partial FKL(p,q_{\theta})}{\partial z^q_j} = q_{\theta}(Y_j|\mathbf{y}_{<t}) - p(Y_j|\mathbf{y}_{<t}) \ \forall j \in \{1,2,3,...,V\},
\end{equation}
and then
\begin{equation*}
\frac{\partial FKL(p,q_{\theta})}{\partial z^q_j} = 0 \ \forall j \in \{1,2,3,...,V\} \Leftrightarrow q_{\theta}(Y_j|\mathbf{y}_{<t}) = p(Y_j|\mathbf{y}_{<t}) \ \forall j \in \{1,2,3,...,V\}. \qedhere
\end{equation*}
\end{proof}

\begin{proof}
Considering the converge condition for RKL:
\begin{equation}
\frac{\partial RKL(p,q_{\theta})}{\partial z^q_j} = 0 \ \forall j \in \{1,2,3,...,V\}.
\label{rg}
\end{equation}
We have
\begin{equation}
\frac{\partial RKL(p,q_{\theta})}{\partial z^q_j} = q_{\theta}(Y_j|\mathbf{y}_{<t})(\log(\frac{ q_{\theta}(Y_j|\mathbf{y}_{<t})}{p(Y_j|\mathbf{y}_{<t})})-RKL(p,q_{\theta})).
\label{rkl_g}
\end{equation}
\textit{Sufficiency.} \ Put $q_{\theta}(Y_j|\mathbf{y}_{<t}) = p(Y_j|\mathbf{y}_{<t}) \ \forall j \in \{1,2,3,...,V\}$ into \ref{rkl_g}:
\begin{equation}
\begin{aligned}
\frac{\partial RKL(p,q_{\theta})}{\partial z^q_j} &= q_{\theta}(Y_j|\mathbf{y}_{<t})(\log(\frac{ q_{\theta}(Y_j|\mathbf{y}_{<t})}{p(Y_j|\mathbf{y}_{<t})})-RKL(p,q_{\theta})) \\
&=q_{\theta}(Y_j|\mathbf{y}_{<t})RKL(p,q_{\theta}) \\
&=q_{\theta}(Y_j|\mathbf{y}_{<t})(
\sum\limits_{m=1}^{V}
q_{\theta}(Y_m|\mathbf{y}_{<t})
\log(\frac{q_{\theta}(Y_m|\mathbf{y}_{<t})}{p(Y_m|\mathbf{y}_{<t})})
) \\
&= 0 \ \forall j \in \{1,2,3,...,V\}
\end{aligned}
\end{equation}
\textit{Necessity.} When Equation \ref{rg} holds, then $\forall i,j \in \{1,2,3,...,V\}$, we can get:
\begin{equation}
q_{\theta}(Y_i|\mathbf{y}_{<t})
\frac{\partial RKL(p,q_{\theta})}{\partial z^q_j} - q_{\theta}(Y_j|\mathbf{y}_{<t})
\frac{\partial RKL(p,q_{\theta})}{\partial z^q_i} = q_{\theta}(Y_i|\mathbf{y}_{<t}) * 0 - q_{\theta}(Y_j|\mathbf{y}_{<t}) * 0 = 0.
\label{qsum}
\end{equation}
Put Equation \ref{rkl_g} into Equation \ref{qsum}, thus
\begin{equation}
q_{\theta}(Y_i|\mathbf{y}_{<t})q_{\theta}(Y_j|\mathbf{y}_{<t})
\log(\frac{ q_{\theta}(Y_j|\mathbf{y}_{<t})p(Y_i|\mathbf{y}_{<t})}{p(Y_j|\mathbf{y}_{<t})q_{\theta}(Y_i|\mathbf{y}_{<t})})=0.
\end{equation}
Meanwhile, we have $0<q_{\theta}(Y_i|\mathbf{y}_{<t})<1$ and $0<q_{\theta}(Y_j|\mathbf{y}_{<t})<1$, therefore
\begin{equation}
\begin{aligned}
\frac{q_{\theta}(Y_j|\mathbf{y}_{<t})}{q_{\theta}(Y_i|\mathbf{y}_{<t})} &= \frac{p(Y_j|\mathbf{y}_{<t})}{p(Y_i|\mathbf{y}_{<t})}
\ \forall i,j \in \{1,2,3,...,V\}
\\
\rightarrow 
\sum\limits^V_{i=1}
\frac{q_{\theta}(Y_i|\mathbf{y}_{<t})}{q_{\theta}(Y_m|\mathbf{y}_{<t})}&=
\sum\limits^V_{i=1}
\frac{p(Y_i|\mathbf{y}_{<t})}{p(Y_m|\mathbf{y}_{<t})} \ \forall m \in \{1,2,3,...,V\}
\\
\rightarrow 
\frac{\sum\limits^V_{i=1} q_{\theta}(Y_i|\mathbf{y}_{<t})}{q_{\theta}(Y_m|\mathbf{y}_{<t})}&=
\frac{\sum\limits^V_{i=1}p(Y_i|\mathbf{y}_{<t})}{p(Y_m|\mathbf{y}_{<t})} \ \forall m \in \{1,2,3,...,V\}
\\
\rightarrow 
\frac{1}{q_{\theta}(Y_m|\mathbf{y}_{<t})}&=
\frac{1}{p(Y_m|\mathbf{y}_{<t})} \ \forall m \in \{1,2,3,...,V\}
\\
\rightarrow 
q_{\theta}(Y_m|\mathbf{y}_{<t})&=
p(Y_m|\mathbf{y}_{<t}) \ \forall m \in \{1,2,3,...,V\}.
\end{aligned}
\end{equation}
To summarize, we have:
\begin{equation*}
\frac{\partial RKL(p,q_{\theta})}{\partial z^q_j} = 0 \ \forall j \in \{1,2,3,...,V\} \Leftrightarrow q_{\theta}(Y_j|\mathbf{y}_{<t}) = p(Y_j|\mathbf{y}_{<t}) \ \forall j \in \{1,2,3,...,V\}. \qedhere
\end{equation*}
\end{proof}

\section{Evaluation details}
\label{appendix: dataset_detail}
The details for the evaluation dataset are as follows:
\begin{itemize}
    \item Dolly: human-written instruction-response pairs \footnote{https://github.com/databrickslabs/dolly/tree/master}.
    We divide the data set into 14k samples for training, 500 samples for validation, and 500 samples for testing following \citet{DBLP:journals/corr/abs-2306-08543}.
    \item S-NI: the test set of SUP-NATINST \citep{wang-etal-2022-super}, which contains 9K samples from 119 English tasks.
    In this paper, we employ the samples with ground truth responses longer than 11.
    \item UnNI: dataset from \citet{honovich-etal-2023-unnatural}.
    Similarly, we employ the samples with ground-truth responses longer than 11.
\end{itemize}

We use five random seeds $\{10, 20, 30, 40, 50\}$ for generation.

\section{More Baselines}
\label{appendix: more_baseline}

In Table \ref{main_res}, we report a baseline FKL+RKL defined as 0.5FKL+0.5RKL.
For more baselines, we set the loss weights as 0.25/0.75 and 0.75/0.25, respectively. 
Table \ref{table: more_baseline} shows the results and we can find that AKL outperforms all the baselines.

\begin{table}[h]
\centering
\resizebox{0.6\linewidth}{!}
{%
\begin{tabular}{lccc}
\toprule
\textbf{Methods} & \textbf{Dolly} & \textbf{S-NI} & \textbf{UnNI} \\
\midrule
0.50 FKL + 0.50 RKL & 23.36$\pm$0.53 & 17.83$\pm$0.23 & 20.37$\pm$0.08 \\
0.25 FKL + 0.75 RKL & 22.53$\pm$0.48 & 16.18$\pm$0.23 & 19.28$\pm$0.11 \\
0.75 FKL + 0.25 RKL & 23.56$\pm$0.37 & 17.76$\pm$0.40 & 19.17$\pm$0.13 \\
\rowcolor{gg} AKL~(Ours) & 23.88$\pm$0.46 & 19.15$\pm$0.21 & 21.97$\pm$0.13 \\
\bottomrule
\end{tabular}
}
\caption{
Results of more baselines from GPT2 1.5B teacher to GPT2 120M student.
}
\label{table: more_baseline}

\end{table}

\section{BERTscore Results}
\label{appendix: bert score}

To evaluate the generated responses, we report the rouge scores in Table \ref{main_res}, GPT-4 scores in Figure \ref{gpt4_scores}, and human evaluation results in Table \ref{table: human_evalu}. 
Nevertheless, we also calculate the F1 BERTscore \citep{DBLP:conf/iclr/ZhangKWWA20}.
Table \ref{table: bert_score} reports the BERTscore on the generated responses from TinyLLaMA.
AKL consistently outperforms all the baselines on the BERTscore. 
Moreover, the p-values for the conclusion that AKL outperforms FKL+RKL are less than 0.05~(0.002, 0.028, and 0.00006, respectively).

\begin{table}[h]
\centering
\resizebox{0.55\linewidth}{!}
{%
\begin{tabular}{lccc}
\toprule
\textbf{Methods} & \textbf{Dolly} & \textbf{S-NI} & \textbf{UnNI} \\
\midrule
Teacher &  86.55$\pm$0.18 & 87.27$\pm$0.03 & 86.32$\pm$0.01 \\
SFT	 & 85.45$\pm$0.16 & 86.06$\pm$0.04 & 84.91$\pm$0.03 \\FKL &   85.40$\pm$0.16 & 86.15$\pm$0.12 & 85.07$\pm$0.02 \\
RKL  &    85.92$\pm$0.07  &
86.46$\pm$0.08 & 85.48$\pm$0.04 \\
FKL+RKL &  85.62$\pm$0.15  & 86.82$\pm$0.09 & 85.75$\pm$0.01 \\
SeqKD & 85.48$\pm$0.18  & 85.19$\pm$0.08 & 85.04$\pm$0.02 \\
SKL & 85.72$\pm$0.18 & 86.80$\pm$0.05 & 85.48$\pm$0.02 \\
SRKL & 85.57$\pm$0.07 & 86.55$\pm$0.08 & 85.40$\pm$0.03 \\
\rowcolor{gg} AKL (ours) &  86.01$\pm$0.14 & 86.92$\pm$0.04 & 85.89$\pm$0.04  \\
\bottomrule
\end{tabular}
}
\caption{
Results of BERTscores on the generated responses from TinyLLaMA.
}
\label{table: bert_score}

\end{table}

\section{Sensitivity of $\mu$ in AKL}
\label{appendix: akl_mu_sens}

$\mu$ is a threshold to separate the head/tail parts. Since the sum of all logits is 1, thus we intuitively set $\mu$ as 0.5 for head/tail split.
Table \ref{table: mu_sense} shows the results of more values for $\mu$.
We can find that AKL with $\mu$=0.45/0.55 would also outperform FKL+RKL, and even better than AKL ($\mu$=0.5), showing the robustness of the proposed AKL.

\begin{table}[h]
\centering
\resizebox{0.6\linewidth}{!}
{%
\begin{tabular}{lccc}
\toprule
\textbf{Methods} & \textbf{Dolly} & \textbf{S-NI} & \textbf{UnNI} \\
\midrule
FKL+RKL       & 24.08$\pm$0.51 & 30.98$\pm$0.31 & 30.48$\pm$0.10 \\
AKL ($\mu$=0.50) & 24.40$\pm$0.42 & 31.37$\pm$0.23 & 31.05$\pm$0.17 \\
AKL ($\mu$=0.45) & 24.45$\pm$0.14 & 31.69$\pm$0.25 & 32.48$\pm$0.08 \\
AKL ($\mu$=0.55) & 24.29$\pm$0.53 & 31.79$\pm$0.36 & 32.38$\pm$0.06 \\
\bottomrule
\end{tabular}
}
\caption{
Comparison of results for different $\mu$ in AKL from LLaMA 6.7B to TinyLLaMA 1.1B.
}
\label{table: mu_sense}

\end{table}

\section{Scores from GPT-4}
\label{appendix: gpt4score}
We design prompts to get scores from the GPT-4 model.
For the diversity score, the prompt is: 
\begin{boxA}
Give a score (1-10, the higher the better) to judge the diversity of 5 answers to one question. 
Q: [\textit{Input}] 
Ans 1: [\textit{Answer 1}] 
Ans 2: [\textit{Answer 2}] 
Ans 3: [\textit{Answer 3}]  
Ans 4: [\textit{Answer 4}] 
Ans 5: [\textit{Answer 5}].
Just output the score.
\end{boxA}
where \textit{Input} and \textit{Answer 1-5} denote the corresponding input and responses under five seeds~(10, 20, 30, 40, 50).
For the quality score, we first get the score for each answer and then calculate the average, aiming to avoid the influence of other answers.
The prompt is designed as:
\begin{boxA}
Give a score~(1-10, the higher the better) to judge the quality of the answer to the question. 
Q: [\textit{Input}] Ans: [\textit{Answer}].
Just output the score.
\end{boxA}
For the five responses from the same input, we record the individual scores and calculate the average as the final score.
Due to the cost, we employ the responses from the Open QA and Closed QA subtasks.

\section{Results on more benchmarks}
\label{appendix: more_benchmark}
We further evaluate the performance of TinyLLaMa on the following benchmarks:
\begin{itemize}
    \item Winogrande \citep{sakaguchi2021winogrande}: the task of common sense reasoning, we report the accuracy under the setting of zero-shot learning;
    \item OpenBookQA \citep{mihaylov2018can}: question answering task modeled after open book exams, we report the accuracy under the setting of zero-shot learning;
    \item BoolQ \citep{clark2019boolq}:  question answering task for yes/no questions, we report the accuracy under the setting of zero-shot learning;  
    \item ARC \citep{clark2018think}: a new dataset of 7787 genuine grade-school level, multiple-choice science questions, we report the normalized accuracy of the challenge part under the 5-shot prompt.
\end{itemize}

We implement the evaluation based on the LM-Evaluation-Harness \citep{eval-harness} framework.
As shown in Table \ref{more_benchmark}, our proposed AKL outperforms all the baselines, which indicates the robustness of AKL.

\begin{table}[h]
\centering
\resizebox{0.6\columnwidth}{!}
{%
\begin{tabular}{lcccc}
\toprule
\textbf{Methods} & \textbf{Winogrande} & \textbf{OpenBookQA} & \textbf{BoolQ} & \textbf{ARC} \\
\midrule
Teacher & 68.70 & 37.20 & 78.56 & 51.71 \\
SFT & 59.03 & 27.60 & 62.50 & 35.84 \\
\midrule
SeqKD & 59.11 & 26.80 & 62.23 & 36.43 \\
FKL & 59.27 & 26.80 & 63.14 & 36.09 \\
RKL & 59.66 & 27.80 & 63.45 & 36.09 \\
FKL+RKL & 59.58 & 28.20 & 62.99 & 35.92 \\
\rowcolor{gg}AKL~(Ours) & \textbf{60.77} & \textbf{28.60} & \textbf{63.48} & \textbf{36.52} \\
\bottomrule
\end{tabular}
}
\caption{The scores of TinyLLama on more benchmarks.
\textbf{Bold} denotes the best.}
\label{more_benchmark}

\end{table}

\section{Performance on Subtasks}
\label{appendix: rouge_subdomain}
\begin{table*}[t]
\centering
\resizebox{0.95\linewidth}{!}
{%
\begin{tabular}{lcccccccc}
\toprule

\textbf{\multirow{2}{*}{Methods}} & \textbf{Closed} & \textbf{General} & \textbf{Open} & \textbf{Information}  & \textbf{\multirow{2}{*}{Summary}} & \textbf{Brain} & \textbf{Creative}   & \textbf{\multirow{2}{*}{Classification}} \\
 & \textbf{QA} & \textbf{QA} & \textbf{QA} & \textbf{Extraction} & & \textbf{Storming} & \textbf{Writing} & \\
\midrule
Teacher & 31.52 & 16.95 & 24.99 & 24.60 & 30.73 & 23.21 & 14.62 & 54.12 \\
SFT & 22.75&	14.43&	16.68&	22.09&	25.21&	16.27&	13.48&	46.64 \\
\midrule
SeqKD & 22.95&	15.51&	17.47&	24.68 &	27.27&	15.79&	13.52&	\underline{51.60} \\
FKL & \textbf{27.46}&	16.07&	17.68&	24.30&	\textbf{28.10}&	15.59&	14.15&	47.01 \\
RKL & 23.99&	\underline{16.32}&	\underline{20.07}&	19.54&	26.27&	\textbf{19.40}&	13.76&	51.39 \\
SKL & 24.32 & 15.66 & 18.51 & 25.29 & 26.47 & 17.30 & 14.15 & 45.93 \\
SRKL& 23.77 & 16.20 & 17.05 & \textbf{25.45} & 24.61 & 16.23 & 14.04 & 46.11 \\
FKL+RKL &24.49&	14.94&	18.40&	21.79&	25.33&	17.13&	\underline{14.29} &	48.00 \\
\rowcolor{gg}AKL~(Ours) &\underline{26.85}&	\textbf{16.79}&	\textbf{21.17}&	\underline{25.43}&	\underline{27.55}&	\underline{18.69}&	\textbf{15.00}&	\textbf{53.19} \\
\bottomrule
\end{tabular}
}
\caption{Rouge-L scores for TinyLLaMA 1.1B on different subtasks from Dolly.
\textbf{Bold} denotes the best results and \underline{underline} for the second.
}
\label{rouge_sta}
\end{table*}
In Table \ref{rouge_sta}, we showcase the detailed performance regarding 8 subtasks from Dolly. 
FKL and RKL perform well on different sub-tasks~(i.e. FKL on Closed QA and RKL on Brain Storming).
Meanwhile, our proposed AKL can combine the advantages of FKL and RKL well while FKL+RKL fails.

\section{Cases}
\label{appendix: case}
We show some cases for the outputs of baselines and proposed AKL.
AKL can effectively inherit the knowledge from the teacher model, such as the artistry~(see Table \ref{case1}), grammar information~(see Table \ref{case4}), correct information~(see Table \ref{case3}), and representation skill~(see Table \ref{case2}).

\begin{table}[ht]
\begin{tabular}{lp{0.8\textwidth}}
\toprule
\textbf{Input} & \textbf{Write a haiku about how much I love pickles.} \\
\midrule
GPT-4 & Crunchy, briny treat,/ Love for pickles knows no bounds, / 
Savor every bite.
\\
\midrule
Teacher &  Pickles I love,  / 
Like a salty hug.
 \\
\midrule
SFT &  Syrup so sweet, pickles my soul.
\\
\midrule
SeqKD & A slipping pickle,/ 
With melty mayo drizzled high,/ 
Aha, divine!
 \\
\midrule
FKL & Salty goodness in a bundle... /
I love pickles!
 \\
\midrule
RKL &  I love pickles so much/
They're definitely delicious/
You should try them too
 \\
\midrule
SKL & So beautiful in May / I'll take a bite of yours / Every day are as sweet as pickles. \\
\midrule
SRKL & Tasty pickles keep me fueled. / Making them takes nerve, / But be still my pickle-headed soul!\\
\midrule
FKL+RKL &  Love is fickle, /
But pickles are eternal.
 \\
\midrule
AKL~(Ours) & Bulbs of great deliciousness /
Shining in the sun /
Like precious jewels.
 \\
\bottomrule
\end{tabular}
\caption{One case from the \textbf{Creative Writing} subtask in Dolly.
Only AKL and FKL+RKL learn the artistry from the teacher model.}
\label{case1}
\end{table}
\begin{table}[ht]
\begin{tabular}{lp{0.8\textwidth}}
\toprule
\textbf{Input} & \textbf{Below is an instruction that describes a task, paired with an input that provides further context. Write a response that appropriately completes the request. \text{\#\#\#}Instruction: Given a description of an object, answer the question 'What is it?' \text{\#\#\#}Input: A thing that has mass and occupies space.} \\
\midrule
GPT-4 & It is matter.
\\
\midrule
Teacher &  A thing that has mass and occupies space is call \textbf{an object}.
 \\
\midrule
SFT &  It is \textbf{a object}!
\\
\midrule
SeqKD &  a glove
 \\
\midrule
FKL &   It is exactly what it seems - a statement that is exactly true. It is a mass/space-occupying object, and much more.
 \\
\midrule
RKL &  A thing that Has Mass and Occupies Space
 \\
\midrule
SKL & A thing that is shape-able with the aid of tools is commonly referred to as an object. \\
\midrule
SRKL & A thing that is shape-able with the aid of tools into a specific shape. Sometimes called a 'mould'. \\
\midrule
FKL+RKL &  A thing that has mass and occupies space
 \\
\midrule
AKL~(Ours) &  \textbf{An object} is a thing that has mass and occupies space.
 \\
\bottomrule
\end{tabular}
\caption{One case from the \textbf{UnNI} dataset.
There is a grammar mistake for the student model~(SFT, 'a' should be 'an').
AKL corrects that mistake with the knowledge from the teacher model.}
\label{case4}
\end{table}
\begin{table}[ht]
\begin{tabular}{lp{0.8\textwidth}}
\toprule
\textbf{Input} & \textbf{Tell me whether these cities are in Texas: Austin, Houston, New York, Chicago, Miami, Dallas.
} \\
\midrule
GPT-4 & In Texas, you will find the following cities from your list: 1. Austin; 2. Houston; 3. Dallas. The other cities you mentioned are located in different states: 1. New York is in New York State; 2. Chicago is in Illinois; 3. Miami is in Florida.
 \\
\midrule
Teacher &  Austin, Houston and Dallas are all cities in Texas.
New York, Chicago and Miami are cities in other states.
However, there is a Miami in Texas as well! 
Texas Miami is near the Florida one, but is still a completely different city.
\\
\midrule
SFT &  Austin and Dallas are in Texas. Everyone else is not in Texas.
\\
\midrule
SeqKD &  Austin and Houston are in Texas.  New York, Chicago, and Dallas are not in Texas.  Dallas is in the state of Texas, while Texas is not alone in the United States.  Miami is located in Florida, and Houston is part of the state of Texas, but is located outside of the American Samoa territory.  American Samoa is the only territory within the 50 United States that is not owned by the United States.  The last city, Texas, does not belong to any of those listed.
 \\
\midrule
FKL &  Austin and Dallas are in Texas.  Houston, New York and Chicago are cities outside of Texas.
 \\
\midrule
RKL &  Austin. Houston. Dallas. New York. Chicago. Miami.
\\
\midrule
SKL & Austin and Dallas are in Texas. The rest are not. \\
\midrule
SRKL & Austin, Houston, Dallas are in Texas. New York, Chicago, and Miami are not in Texas \\
\midrule
FKL+RKL & Austin, Houston, and Dallas are in Texas. New York, Chicago, and Miami are Not in Texas. \\
\midrule
AKL~(Ours) &  Austin, Houston, and Dallas are in Texas. New York and Chicago are not in Texas. Miami is in Florida, and Dallas is in Texas.
 \\
\bottomrule
\end{tabular}
\caption{One case from the \textbf{Classification} subtask in Dolly.
AKL is the only method that both 1) get the correct answer and 2) figure out that Miami is in Florida.}
\label{case3}
\end{table}
\begin{table}[ht]
\begin{tabular}{lp{0.83\textwidth}}
\toprule
\textbf{Input} & \textbf{What should I do on a free afternoon in San Francisco?} \\
\midrule
Teacher &  There are many options for someone with an afternoon free in San Francisco. The city has a wide variety of museums, shops, and landmarks. The Golden Gate Bridge is a must-see if you haven't seen it yet. If museums are your thing, then the De Young and the Legion of honor are both fine options as well. One interesting lesser known option is to walk across the market near the Ferry Building and then walk along the Embarcadero until you get to the Oakland Bay Bridge. Then, as you head back, stop into the Rockridge district for some great wine and cheese at shops like Bioulet, and shop for groceries at the Rasputin deli. If none of these strike your fancy, then San Francisco has a number of taxi cabs and Uber/Lyft rides will be very affordable and easy to come by. \\
\midrule
SFT &  When in San Francisco, you should start with a walk down Market Street. Gridley Gallery and the Legion of Honor are also recommended. If you like shopping, try Union Square. Find a map located at the info booths around the city and figure out which tour boats go down the Embarcadero. If you like taking pictures, then head up to the Fort Mason area, specifically the Canada Gallery. Have some dinner at Japanese forest and then walk Little H influencer. End your evening at the standard Pier 39.
\\
\midrule
SeqKD &  Find a nice restaurant to eat at, take a walk along the Embarcadero, browse at various boutique as you walk along, then win at the casino at the Fisher Fine Arts Museum.
 \\
\midrule
FKL & San Francisco is a very walkable city, so there are many free attractions in San Francisco. Visitors preferably take a trolley tour to see all the main attractions, but walkable as well.
Here are some free attractions: Golden Gate Park; DeYoung Museum; Golden Gate Bridge; Museum of Contemporary Irish and International Art; de Young Museum; Golden Gate Park parade; Explorie trail/mile long iterpretive park trail through Golden Gate Park.
 \\
\midrule
RKL & Go to the DeYoung Museum, pick up lunch at K \& L Bakery on Polk Street, and walk around Chestnut Street Studio District.
\\
\midrule
SKL & There are many free activities in San Francisco. You can visit the Golden Gate Park, which is free to enter. You can also visit the deYoung Museum, the lillie museum, or the Legion of Honor. If you are an art enthusiast, these museums are free. If you prefer viewing artworks that are outdoors, you can walk around the Reflective Pool, Pavilion, or Rocky Point. The view of the bay is beautiful during the day, so you can also visit the Golden Gate Bridge, which is also free. \\
\midrule
SRKL & There are a lot of historical buildings in Golden Gate Park, and you can visit any of them free of charge. For example, the DeYoung Museum, the Civic Center Branch of the Library, the Museum of Modern Art, and the Tennis Museum are all in Golden Gate Park. \\
\midrule
FKL+RKL & One could explore Golden Gate Park and the  SF Museum of Modern Art, or simply walk around the city. The Italian Street Art walk is also fun. \\
\midrule
AKL~(Ours) &   There are many things to do and see in San Francisco. You can visit the more famous landmarks such as Lombard Street cycle and the Golden Gate Bridge. Downtown San Francisco is also home to numerous restaurants and bars. If you are interested in history, then the Golden Gate Park is a great place to spend a afternoon exploring the local art scene. You can also walk around the Fisher House, Hilltop Park and the Conservatory of Flowers. For the more adventurous, there are many walking tours of San Francisco, including the trolley tour and one of the Vessel San Francisco.
 \\
\bottomrule
\end{tabular}
\caption{One case from the \textbf{Creative Writing} subtask in Dolly.
Only AKL maintains the style~(i.e. if you are interested in \#\#, you can \#\#) and generates rich suggestions.}
\label{case2}
\end{table}

\end{document}